\documentclass[conference]{IEEEtran}
\IEEEoverridecommandlockouts
\usepackage{epsfig,makeidx,color}
\usepackage{amsmath,amssymb,bbm}
\usepackage{cite,graphicx,lipsum}
\usepackage{enumerate,balance}
\usepackage[switch,pagewise]{lineno}
\usepackage{hyperref}
\usepackage{caption}
\usepackage{subcaption}
\usepackage{soul}
\usepackage[normalem]{ulem} 
\hypersetup{
        colorlinks = true,
        citecolor=blue,
}
\def\cG{{\cal G}}

\def\rT{{\rm T}}

\def\uB{{\mathbb B}}

\DeclareMathOperator*{\argmin}{\arg\!\min}

\def\be{ \begin{equation} }
\def\ee{ \end{equation} }
\def\bea{ \begin{eqnarray} }
\def\eea{ \end{eqnarray} }

\def\bd{{\bf d}}

\def\ba{{\bf a}}

\def\bee{{\bf e}}

\def\bone{{\bf 1}}

\def\b0{{\bf 0}}

\def\cA{{\cal A}}

\def\cP{{\cal P}}
\def\cO{{\cal O}}

\ifCLASSOPTIONonecolumn
\else

\fi

\title{Energy-Efficient Downlink Semantic Generative Communication with Text-to-Image Generators}

\author{\IEEEauthorblockN{Hyein Lee\IEEEauthorrefmark{1},
Jihong Park\IEEEauthorrefmark{2}, Sooyoung Kim\IEEEauthorrefmark{1}, and
Jinho Choi\IEEEauthorrefmark{2}}
\IEEEauthorblockA{\IEEEauthorrefmark{1}IT Convergence Research
Center,
Division of EE, Jeonbuk National University, 
Email: \{leehyein96,sookim\}@jbnu.ac.kr} 
\IEEEauthorrefmark{2}School of IT, Deakin University, Email: {\{jihong.park, jinho.choi\}@deakin.edu.au}
\thanks{This work was supported by the National Research Foundation (NRF) of Korea grant funded by the Korea government (MSIT) (No NRF-2021R1A2C1003121). J. Park, S. Kim, and J. Choi are corresponding authors.}
}

\begin{document}

\maketitle

\begin{abstract}
In this paper, we introduce a novel semantic generative communication (SGC) framework, where generative users leverage text-to-image (T2I) generators to create images locally from downloaded text prompts, while non-generative users directly download images from a base station (BS). Although generative users help reduce downlink transmission energy at the BS, they consume additional energy for image generation and for uploading their generator state information (GSI). We formulate the problem of minimizing the total energy consumption of the BS and the users, and devise a generative user selection algorithm. Simulation results corroborate that our proposed algorithm reduces total energy by up to 54\% compared to a baseline with all non-generative users. 


\end{abstract}
\ifCLASSOPTIONonecolumn
\baselineskip 26pt
\fi

\section{Introduction}

Semantic communication (SC)~\cite{gunduz2022beyond} and generative artificial intelligence (AI) \cite{deepai,openai_dalle,rombach2022high} are two cutting-edge technologies that can create a synergetic effect. Traditional communication systems are concerned with delivering source data in bits, while SC focuses on conveying the meanings (i.e., semantics) of the data. As semantics, such as the text embeddings or topological information of images, are typically smaller and more robust to distortions, SC can greatly reduce communication latency without compromising reliability \cite{maatouk2022age,Jinho_2022_SemCom,seo2022semantic,Calvanese_2021_SemCom,NN_SemCom,DeepJSCC}. The success of SC hinges on the ability to decode the delivered semantics, which in turn determines the minimum size of the semantics that can be sent. In this regard, recent advances in generative AI, particularly text-to-image (T2I) generative models like DeepAI \cite{deepai}, DALL-E \cite{openai_dalle}, Stable Diffusion \cite{rombach2022high}, show great promise as they can decode text prompts into large-scale images with high fidelity.

\begin{figure}
    \centering
    \includegraphics[width=.95\columnwidth]{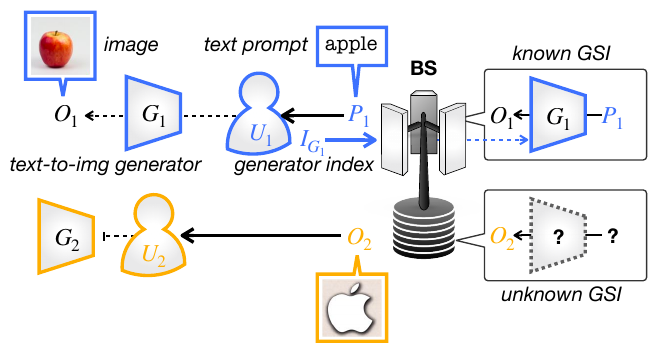}
    \caption{A schematic illustration of downlink semantic generative communication (SGC) with a generative user $U_1$ and a non-generative user $U_2$.}
    \label{Fig:Fig_DS}
\end{figure}

\begin{figure}
    \centering
    \includegraphics[width=\columnwidth]{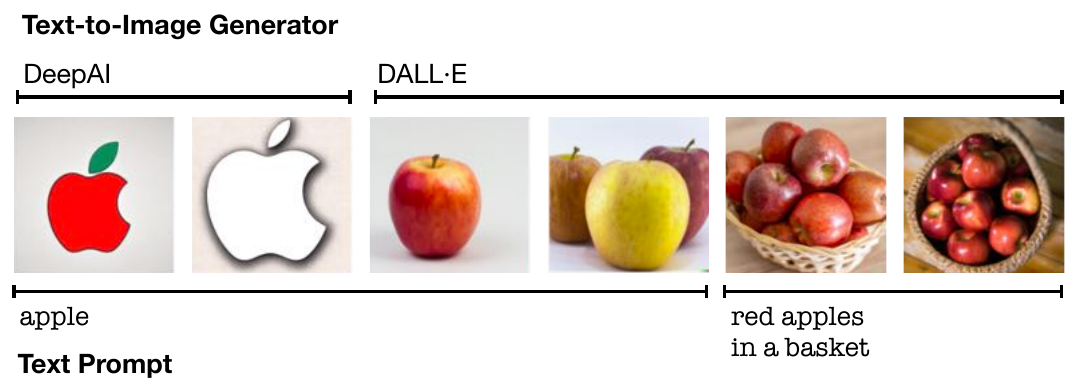}
    \caption{Generated images using the DeepAI and DALL-E text-to-image (T2I) generators.}
    \label{Fig:Fig_Sample}
\end{figure}

Inspired by these developments, we put forward to a novel \emph{semantic generative communication (SGC)} framework for a downlink network, depicted in Fig.~\ref{Fig:Fig_DS}. SGC involves a base station (BS) that aims to deliver images to users ($U_k$) equipped with local T2I generators ($G_k$). These users fall into two categories: \emph{generative users}, who downloads text prompts ($P_k$) to generate images locally, and \emph{non-generative users}, who directly downloading original images ($O_k$). Local T2I generation is non-trivial, resulting in generative users potentially creating non-identical images even when using the same prompt, as demonstrated in Fig.~\ref{Fig:Fig_Sample}. This variability stems from the differences in the users' generators and/or the randomness in their sampling processes. To ensure accurate generation of intended images, each generative user must provide information about its prompt-generator-image mapping ($P_k$-$G_k$-$O_k$), termed \emph{generator state information~(GSI)}. Given that users typically use popular pre-trained generators, we assume that the BS stores all generators, while each generative user only uploads its generator's index $I_{G_k}$ to the BS as GSI.

Within this SGC framework, we formulate the problem of minimizing the total energy consumed by the BS and the users. Generative users consume energy for GSI transmission and image generation, and in return the BS significantly reduces its downlink transmission energy. In contrast, non-generative users do not consume energy, but they increase the downlink transmission energy at the BS. To address this SGC energy minimization problem, we convert it into multiple binary integer linear programming problems, and develop a generative user selection algorithm that considers user-BS channel conditions as well as image and prompt sizes. Simulations confirm that our proposed solution achieves up to 54\% and 46\% reductions in total energy consumption, compared to the baselines with all non-generative users and randomly selected generative users, respectively.

\textbf{Related Works}.\quad
The existing SC frameworks can be broadly classified into three major directions that focus on: task-specific information freshness \cite{maatouk2022age}, knowledge-based communication \cite{Jinho_2022_SemCom,seo2022semantic,Calvanese_2021_SemCom}, and AI-native perceptual compression~\cite{NN_SemCom,DeepJSCC}. SGC is closely aligned with the latter two approaches, wherein generative users function as AI-native receivers, and generators encapsulate their knowledge. Recently, generator model based SC has been explored for distributed metaverse applications \cite{park2022enabling} and AI-native communication \cite{han2022generative}. Notably, in a point-to-point scenario, the latter study utilizes the generative adversarial network (GAN) model mapping Gaussian noise into images, whereas SGC focuses on T2I semantic mappings and multi-user designs. The generative user selection problem in SGC differs from the offloading user selection problem found in mobile computation offloading \cite{Chen16, ko2017MEC,Zhou21}, as the former emphasizes generating computation at users for BS energy reduction, while the latter concentrates on offloading computation from users to a BS.

\section{System Model}

In this section, we describe the system model for downlink SGC, where a single BS intends to send images by transmitting original images to non-generative users or their text prompts to generative users, respectively.

\subsection{Image Generation Task and Data Transmission}

In SGC, we consider a downlink network consisting of 
a BS and $K$ users. 
Suppose that user $k$ has a generative model $G_k$ that can generate 
a set of $L$ images, denoted by $\cO_k = \{O_{k1}, \cdots, O_{kl}, \cdots, O_{kL}\}$, when inputting a text prompt set $\cP_k = \{P_{k1}, \cdots, P_{kl} , \cdots, P_{kL}\}$. 
Let $f(G_k,P_{kl}) = O_{kl}$ denote this image generation task, where  $P_{kl}$ represents a prompt which can produce image $O_{kl}$ 
through user $k$'s generative model, $G_k$. 
For convenience, 
let $C_F$ and $C_k$ denote
the number of central processing unit (CPU) cycles 
to complete $f(G_k,P_{kl})$ and the clock speed of CPU at user $k$, respectively. In addition, $g_k$ stands for the consumed CPU power at user $k$ per second. 
Throughout the paper, we assume that the BS stores not only $\cO_k$ and $\cP_k$, but also all $K$ users' generative models $\cG = \{G_1, \cdots, G_k, \cdots, G_K\}$. 


Suppose that there are two different types of user, i.e., generative and non-generative users.
Let $\ba =[a_1 \ \cdots \ a_K]^\rT$ denote the generative user selection vector, where $a_k \in \{0, 1\}$. Here, if $a_k = 1$, user $k$ transmits its generative model index to the BS so that 
the BS can generate images (this user is called a generative user). Otherwise (i.e., $a_k =0$), user $k$ directly receives desirable images from the BS rather than perform any transmission or computation task (this user is called a non-generative user). In addition, let the index set of generative users be
$$
\cA = \{k: \ a_k = 1, \ k = 1,\ldots, K\}.
$$

For user $k \in \cA$, the following steps are required:
\begin{enumerate}
    \item User $k$ sends an index of generative model it has, $I_{G_k}$, to the BS (via uplink transmission).
\item The BS transmits $P_{kl}$ to user $k$ (via downlink transmission) so that the user $k$ can generate image $O_{kl}$ by itself.

\item User $k$ performs the image generating task $f(G_k,P_{kl})$ using its CPU.
\end{enumerate}
Here, we assume that the computation energy to figure out which $P_{kl}$ creates $O_{kl}$ at the BS is sufficiently small and negligible. On the other hand, for user $k \notin \cA$, the BS has to directly transmit selected images $O_{kl}$. Note that a\textcolor{green}{\sout{n}} generative user needs to have both uplink and downlink transmissions, while a non-generative user needs to have downlink transmission as illustrated in Fig.~\ref{Fig:Fig_DS}.


\subsection{Communication Model}

Throughout the paper, we assume dynamic time division duplexing (TDD) for uplink and downlink transmissions with a time slot that can be flexibly divided.

For uplink transmissions, simultaneous transmissions by multiple generative users are allowed \cite{Chen16}. Let $h_{k}$ denote the channel coefficient between user $k$ to the BS. 
If $k \in \cA$, the uplink transmission rate becomes
\be 
u_k (\ba) = W \log_2 \left( 1 + \frac{\rho_k \beta_k}{
\sum_{i \in \cA \setminus k} \rho_i \beta_i + N_0}
\right),
    \label{EQ:UT}
\ee 
where $W$ is the system bandwidth, $\rho_k$ is the transmit power of user $k$, $\beta_k = |h_k|^2$, and $N_0$ is the variance of the background noise. Thus, signal processing techniques are required to perform multiuser detection at the BS \cite{VerduBook} \cite{ChoiJBook2}.

For downlink transmissions, the BS uses time division multiple access (TDMA) to send data to users. The data rate of downlink transmission to user $k$ is given by
\be 
v_k = W \log_2 \left(1 + 
\frac{\bar \rho_k \beta_k}{N_0}
\right) ,
    \label{EQ:DT}
\ee 
where $\bar \rho_k$ is the transmit power of the BS to user $k$. In \eqref{EQ:DT},
the channel power gain, $\beta_k$, is assumed to be the same as that for uplink transmission because of the channel reciprocity of TDD.
As shown in \eqref{EQ:DT}, each user only needs to perform single-user decoding as there is no interference, while the BS needs to perform multiuser decoding for all generative users as in \eqref{EQ:UT}. 

Note that for uplink transmissions, TDMA can also be used as in \cite{You17}, and a high transmission rate can be achieved with a high transmit power, $\rho_k$. On the other hand, when simultaneous transmissions are considered, due to the interference, the uplink transmission rate in \eqref{EQ:UT_1} becomes limited, while it would not be necessary to have a high transmit power, $\rho_k$. 
Thus, simultaneous transmissions could be desirable for users of limited transmit power.

From \eqref{EQ:UT} and \eqref{EQ:DT},
the downlink transmission time of user $k$ is given by
\begin{align}
\bar T_k & = \left\{
\begin{array}{ll}
\frac{B(P_{kl})}{v_k}, & \mbox{if $k \in \cA$} \cr
  \frac{B(O_{kl})}{v_k}, & \mbox{if $k \notin \cA$},
\end{array}
\right.
\end{align}
where $B(\cdot)$ denotes the size of the data in bits.
For the generative users $k \in \cA$, the uplink transmission time becomes
$T_\text{up} = B(I_{G_k})/u_k (\ba)$ since all the generative users transmit its index of generative model $I_{G_k}$ simultaneously. Moreover, the size of $I_{G_k}$s are the same, i.e., $B(I_{G_k}) = \lceil \text{log}_2(K) \rceil$ (bit) $\forall k$, where $\lceil x \rceil$ represents the least integer greater than or equal to $x$.
As mentioned earlier, the total transmission time of a generative user includes both uplink and downlink transmission time. On the other hand, the total transmission time of a non-generative user only includes downlink transmission time.

\subsection{Energy Model}

For generative user, it requires energy for uplink and downlink transmission as well as computation for the task $f(G_k,P_{kl})$. On the other hand, the case of non-generative user consumes energy only for downlink transmission to transmit images from BS. As a result, we have the total consumed energy as follows:
\begin{align}
E_k & = \left\{
\begin{array}{ll}
g_k \frac{C_F}{C_k} + \rho_k \frac{B(I_{G_k})}{u_k (\ba)} + \bar \rho_k  \frac{B(P_{kl})}{v_k}, & \mbox{if $k \in \cA$} \cr
\bar \rho_k \frac{B(O_{kl})}{v_k}, & \mbox{if $k \notin \cA$}. 
\end{array}
\right. 
    \label{EQ:Ek}
\end{align}
where the terms for generative users (i.e., $k \in \cA$) are the energy for performing the image generation task, uplink transmission and downlink transmission, respectively. Similarly, the total consumed energy for user $k \notin \cA$ is only about downlink transmission.

As TDD is assumed, signal processing techniques can be used to estimate the channel between a user and BS thanks to the channel reciprocity. Then,
the channel inversion power control policy for uplink and downlink transmission is used as follows:
\be 
\rho_k = \frac{\rho}{\beta_k} \ \mbox{and} \ \bar \rho_k = \frac{\bar{\rho}}{{\beta}_k} ,
\ee 
where $\rho$ and $\bar{\rho}$ represent the effective received signal powers at the BS and a user, respectively. Since the maximum transmit power is limited, when $\beta_k$ is too small (due to deep fading), $\rho_k$ or $\bar \rho_k$ can be higher than the maximum transmit power and has to be truncated. However, for convenience, we do not consider any truncation and assume that the users of deep fading (i.e., $\beta_k \leq \epsilon$ for some $\epsilon > 0$) are excluded.

\section{Minimization of Total Energy with Channel Inversion Power Control} 

In this subsection, we consider an optimization problem to minimize the total consumed energy.

\subsection{Minimizing Energy Subject to Transmission Time}

In addition to the channel inversion power control policy,
we assume that in order to support all $K$ users, a time slot of length $\tau$, is allocated. Then, it is necessary that the total transmission time of all $K$ users for uplink and downlink transmissions is limited in order to ensure the channel reciprocity, i.e., $\sum_{k=1}^K \bar T_k + T_{\rm up} \le \tau$. Here, we assume that the computation time at the BS is sufficiently short and ignored in the total transmission time constraint.
We can formulate the following problem to minimize the total consumed energy subject to the total transmission time constraint:
\begin{align}
\hat \ba & = \argmin_{\ba \in \uB^K} \sum_{k=1}^K E_k \cr 
& \mbox{subject to} \ \sum_{k=1}^K \bar T_k +  T_{\rm up} \le \tau,
    \label{EQ:O1}
\end{align}
where $\uB = \{0,1\}$.
An exhaustive search requires a complexity of order $2^K$.
Thus, for a large $K$, it is necessary to find an approach that can find the solution
with a low complexity.

Note that the objective function in \eqref{EQ:O1} is the total consumed energy at the BS as well as $K$ users. This formulation is analogous to the problem in mobile computation offloading such as \cite{You17,Chen16}. In \cite{Chen16}, the objective function is the total energy consumed only by  users. In \cite{Chen16}, the energy consumed for signal transmissions is not included in the objective function. Compared to them, we consider the total energy consumed by both the BS and users, for transmitting either images or prompts at the BS as well as transmitting generator indices and computing T2I generation at the users.




\subsection{Finding the Solution}   \label{SS:Sol}

In this subsection, we show that the problem in \eqref{EQ:O1} can be solved by converting it into multiple binary integer linear programming problems. Each problem can be solved by a standard technique such as the branch and bound method \cite{Papadimitriou98}.

In order to find the solution of \eqref{EQ:O1},
we assume that the number of the generative users is $n$
(i.e., $|\cA|= n$). 
Due to the channel inversion power control, from \eqref{EQ:UT}, we can show that
\begin{align} 
u_k (\ba) & = u(n) \cr 
& = W \log_2 \left( 1 + \frac{\gamma}{
(n-1) \gamma + 1}
\right), \ k \in \cA,
    \label{EQ:UT_1}
\end{align}
where $\gamma = \frac{\rho}{N_0}$ is the 
signal-to-noise ratio (SNR) at the BS (for uplink transmission), and
$v_k = v = W \log_2 (1+ \bar{\gamma})$, 
where $\bar{\gamma} = 
\frac{\bar{\rho}}{N_0}$ is the SNR at a user (for downlink transmission). 
As a result, for $n \in \{1,\ldots, K\}$, we have
\begin{align}
T(n) & = T_{\rm up} + \sum_{k=1}^K \bar T_k \cr 
& = T_0 + \frac{B(I_{G_k})}{u(n)} +\sum_{k=1}^K
\biggl( \underbrace{ \frac{B(O_{kl})-B(P_{kl})}{v}}_{=d_k } \biggl) a_k, 
    \label{EQ:Tn}
\end{align}
where $T_0 = \frac{\sum_{k=1}^K B(O_{kl})}{v}$, which is the total transmission time when there is no generative user (i.e., $T(0) = T_0$). 
Similarly, from \eqref{EQ:Ek}, we can show that
\begin{align}
E (n) & = \sum_{k=1}^K E_k \cr 
& = E_0 + \sum_{k=1}^K e_k (n) a_k,
    \label{EQ:En}
\end{align}
where 
$E_0 = \frac{\bar{\rho} B(O_{kl})}{\beta_k v}$ and 
$e_k (n) = g_k \frac{C_F}{C_k} +
\frac{\rho B(I_{G_k})}{\beta_k u(n)}  +
\frac{\bar{\rho} (B(P_{kl}) - B(O_{kl}))}{\beta_k v}$.
Then, for a given $n$, the optimization problem becomes a binary integer linear programming as follows:
\begin{align}
\hat \ba (n) & =   \argmin_{\ba \in \uB^K} E_0 + \bee^\rT \ba \cr
& \mbox{s.t.} \
\left\{
\begin{array}{l}
\bd^\rT \ba \le \tau - T_0 (n)\cr  
\bone^\rT \ba = n, \cr 
\end{array}
\right.
    \label{EQ:O2}
\end{align}
where $\bee = [e_1 (n) \ \cdots \ e_K (n)]^\rT$,
$\bd = [d_1   \ \cdots \ d_K  ]^\rT$, and 
$\bone = [1 \ \cdots \ 1]^\rT$.
The second constraint is due to the assumption that $|\cA| = n$
or $||\ba||_1 = n$.

We can solve the problem in \eqref{EQ:O2} for given $n \in \{1, \ldots, K-1\}$. Let $\hat E (n)$ denote the minimum energy, which is finite if there exists a feasible solution. Otherwise, $\hat E(n) = \infty$. Note that if $n = 0$ or $K$, there is no need to solve \eqref{EQ:O2} as $\hat \ba (n)$
is $\b0$ or $\bone$, respectively. Then, the solution of \eqref{EQ:O1} becomes 
\be 
\hat \ba = \hat \ba (n^\ast),
\ee 
where $\hat E(n^*) \le \hat E(n)$, $\forall n \ne n^\ast$. Note that if $\hat E(n^\ast) = \infty$, then there is no feasible solution.
If no feasible solution is available, the length of time slot, $\tau$, can be increased.


\begin{figure}
     \centering
         \includegraphics[width=.9\columnwidth]{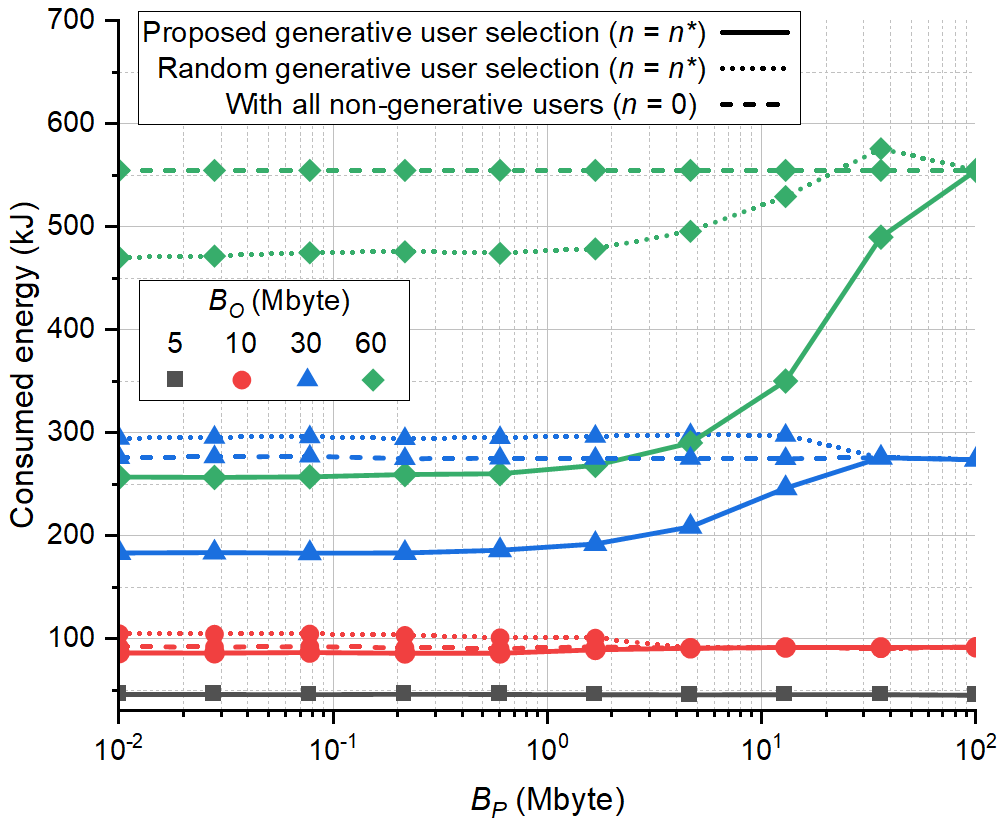}
         \caption{Total consumed energy with with respect to prompt sizes ($B_P$) for different image sizes ($B_O$), with $g_k=45$ W and $\tau=T_0$.}
         \label{Fig:Engergy_1}
\end{figure}

\section{Simulation Results}
The efficiency of the proposed generative user selection scheme is estimated in terms of the consumed energy, by using the simulations with $K=1000$.
We assume that $C_k=3.1$ GHz and $C_F=25\times10^9$ for an image generation task. 
For data transmission, the system bandwidth is $W=5$ MHz and
the SNRs at BS and user are $\gamma=3$ and $\bar{\gamma}=6$, respectively. These SNRs are set from the assumption that the effective received power at user is 2 times higher than the one at BS and the noise condition is not severe. For example, $\rho$, $\bar{\rho}$ and $N_0$ can be 3,6,1 Watt (W), respectively.
In addition, we assume a modified Rayleigh fading such that $\beta_k$ follows a shifted exponential distribution: 
$\beta_k \sim \zeta e^{-\zeta(\beta_k - \epsilon)}$, 
where $\beta_k \ge \epsilon= 0.05$ and $\zeta = \frac{1}{1-\epsilon}$. 
In order to simplify the simulation process, BS is supposed to transmit the same size of prompt and image to generative and non-generative users, respectively, i.e., $B_P = B(P_{kl})$ and $B_O = B(O_{kl})$ $\forall k, l$.
 
\begin{figure}
     \centering
         \includegraphics[width=.9\columnwidth]{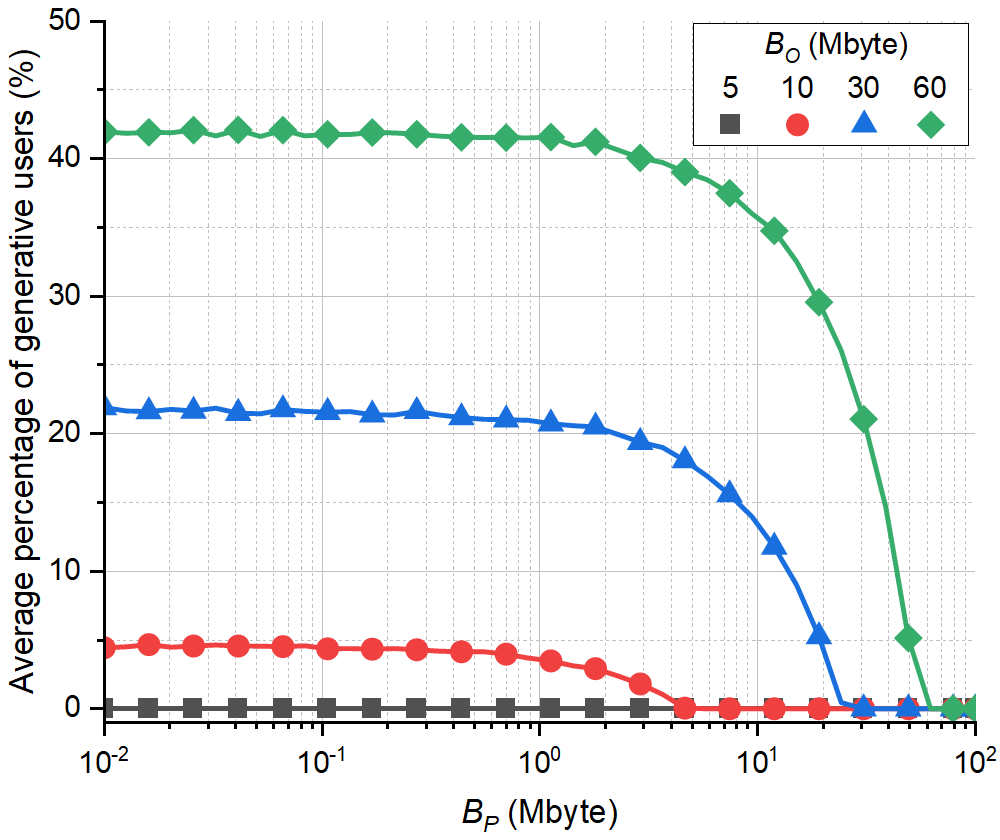}
         \caption{Average ratio of generative users, with respect to prompt sizes ($B_P$) for different image sizes $(B_O)$, with $g_k=45$ W and $\tau=T_0$.}
         \label{Fig:Percentage-of-users_1}
\end{figure}

\begin{figure}
     \centering
     \begin{subfigure}{0.45\columnwidth}
         \centering
         \includegraphics[width=.9\columnwidth]{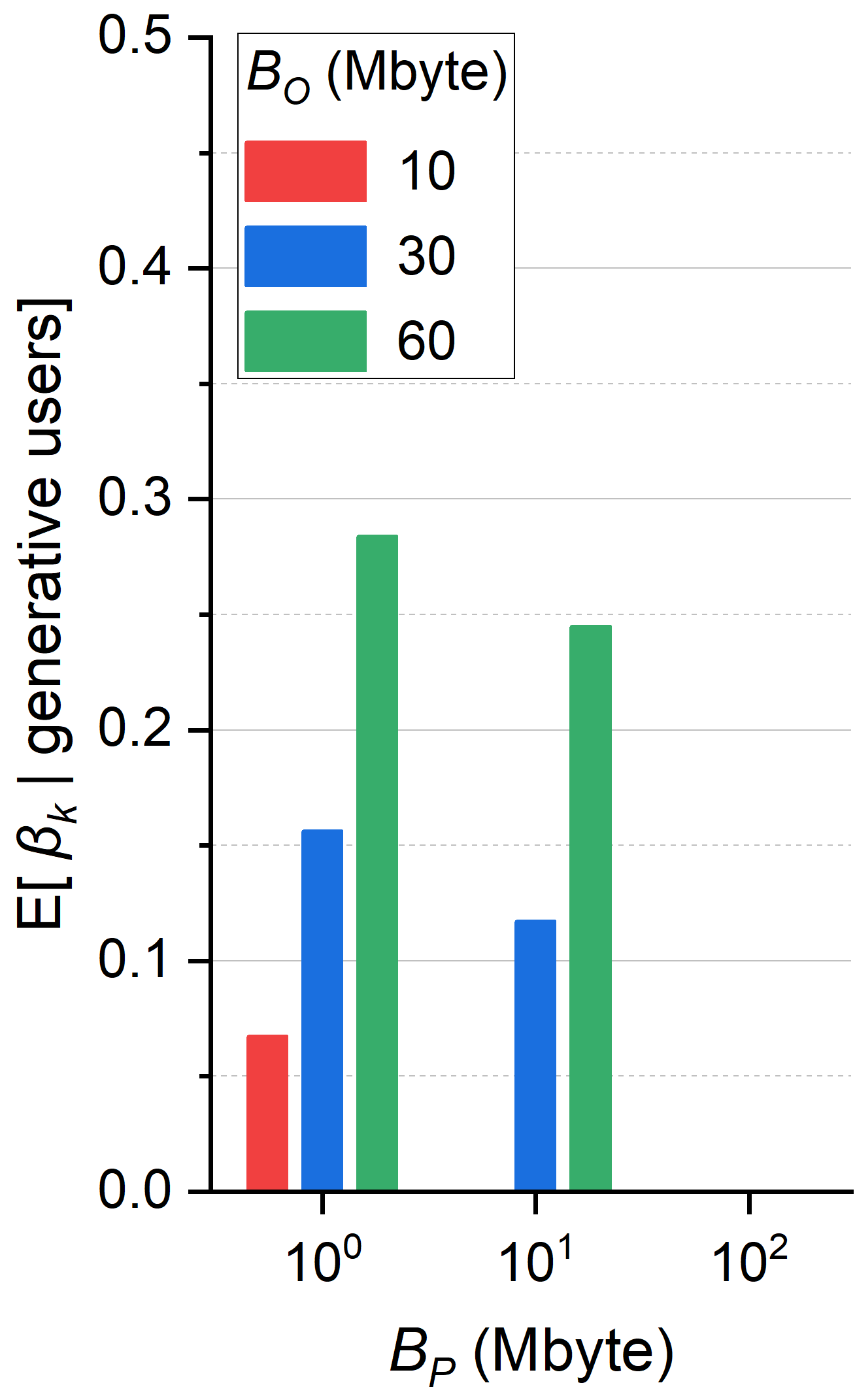}
         \caption{Generative users.}
         \label{Fig:Average_beta_1}
     \end{subfigure}
     \begin{subfigure}{0.45\columnwidth}
         \centering
         \includegraphics[width=.9\columnwidth]{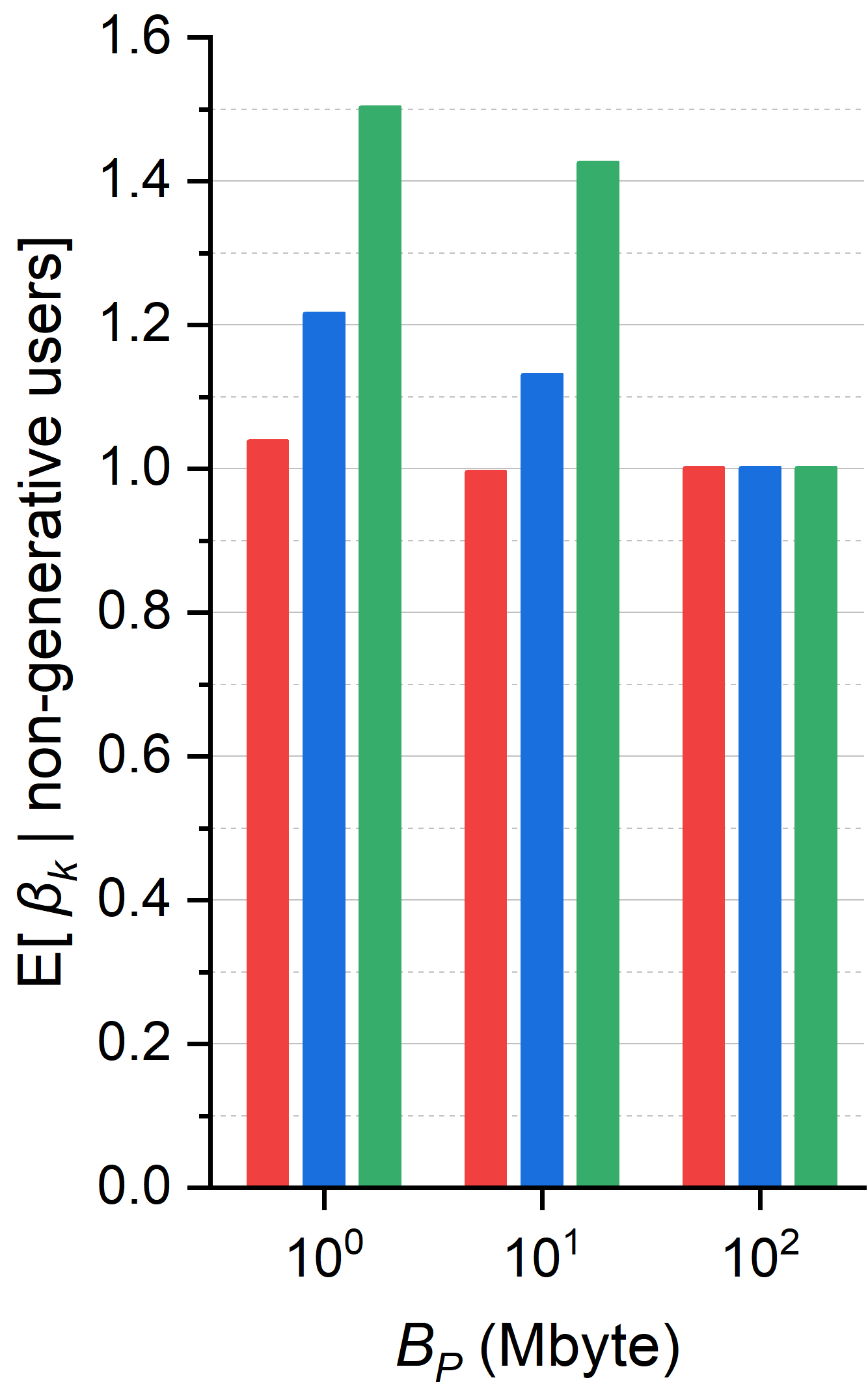}
         \caption{Non-generative users.}
         \label{Fig:Average_beta_2}
     \end{subfigure}
        \caption{Average channel gains (E[$\beta_k$]) for (a) generative users and (b) non-generative users, with respect to different prompt sizes ($B_P$) with $g_k=45$ W and $\tau=T_0$.}
        \label{Fig:Average_beta}
\end{figure}


In Fig.~\ref{Fig:Engergy_1}, we first investigate the consumed energy for various generative situations according to $B_P$ and $B_O$, where $g_k=45$ W and the time constraint threshold $\tau$ is set to the investigated time slot length for $n=K$, i.e., $\tau= T_0$.
Therefore, in this case, BS has sufficient time to transmit images for all $K$ users regardless of the sizes of image.
If there is no generative user, i.e., $n=0$, the consumed energy is proportional to the size of $B_O$. For example, the consumed energy in the range of $B_P<0.2$ Mbyte for $B_O=5,10,30$ and $60$ Mbyte is about 46, 91, 275 and 554 kJ, respectively.
On the other hand, the proposed method shows a strong advantage in energy efficiency as $B_O$ increases,
although its energy consumption is almost the same as the conventional downloading method (i.e., $n=0$) with a small size of image.
For instance, 
the proposed method consumes about 94\% of energy consumed in the conventional scheme when $B_O \leq 10$ Mbyte.
 In contrast, when $B_O$ reaches to 60 Mbyte, the proposed method consumes 
only about 41\% of the energy compared to the case of $n=0$.  
Furthermore, the proposed method for $B_O=60$ Mbyte requires approximately 18\% less energy consumption than transmitting a half-sized image without any generative user (n = 0), i.e., $B_O=30$ Mbyte. Lastly, the proposed generative method is also energy-efficient when the system randomly selects $n=n^*$ generative users with large $B_O$. That means the proposed generative method for large $B_O$ can be applied even when the system is not able to measure channel information.

\begin{figure}
     \centering
         \includegraphics[width=.9\columnwidth]{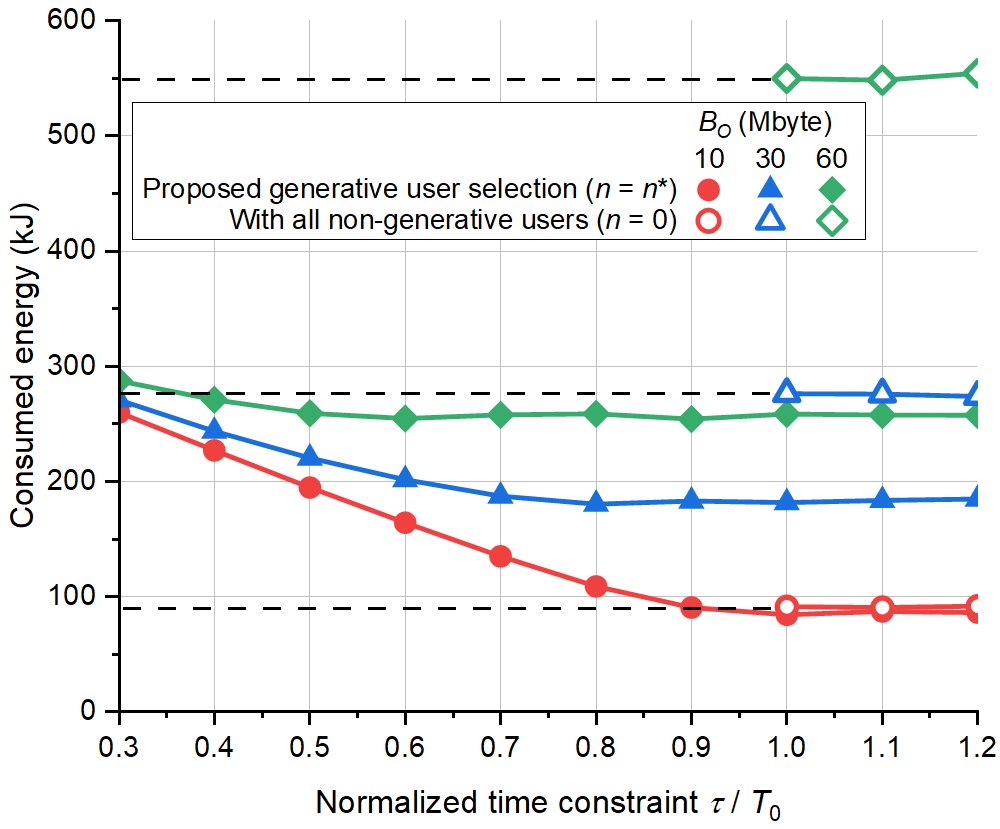}
         \caption{Total consumed energy with respect to different time thresholds ($\tau$), with $g_k=45$ W and $B_P=0.1$ Mbyte, where $T_0$ denotes the total transmission time with all non-generative users.}
         \label{Fig:Energy_2}
\end{figure}

Next, Fig.~\ref{Fig:Percentage-of-users_1} depicts the percentage of generative users according to the sizes of $B_P$ and $B_O$, respectively, where $g_k=45$ W and $\tau= T_0$.
Firstly, it is observed that the size of $B_P$ does not significantly affect 
the number of generative users until it reaches approximately $B_P=0.2$ Mbyte. 
That means, the downlink transmission energy for $B_P$ is negligible compared to the energy consumption required for the computation and the downlink transmission for $B_o$. However, as $B_P$ approaches $B_O$, the energy required for generative users is expended not only for the image generation task but also for the downlink transmission of $B_P$, which is comparable to the energy required for transmission of $B_O$. As a result, the number of generative users converges to 0 before $B_P$ reaches $B_O$.
Unlike $B_P$, a larger $B_O$ increases the percentage of generative users because of
the burden of downlink transmission energy for non-generative users. For instance, when $B_O=60$ Mbyte,  
about 42\% of users are selected as generative users, while it is 0\% for a sufficiently small $B_O$, i.e., $B_O=5$ Mbyte.

\begin{figure}
     \centering
         \includegraphics[width=.9\columnwidth]{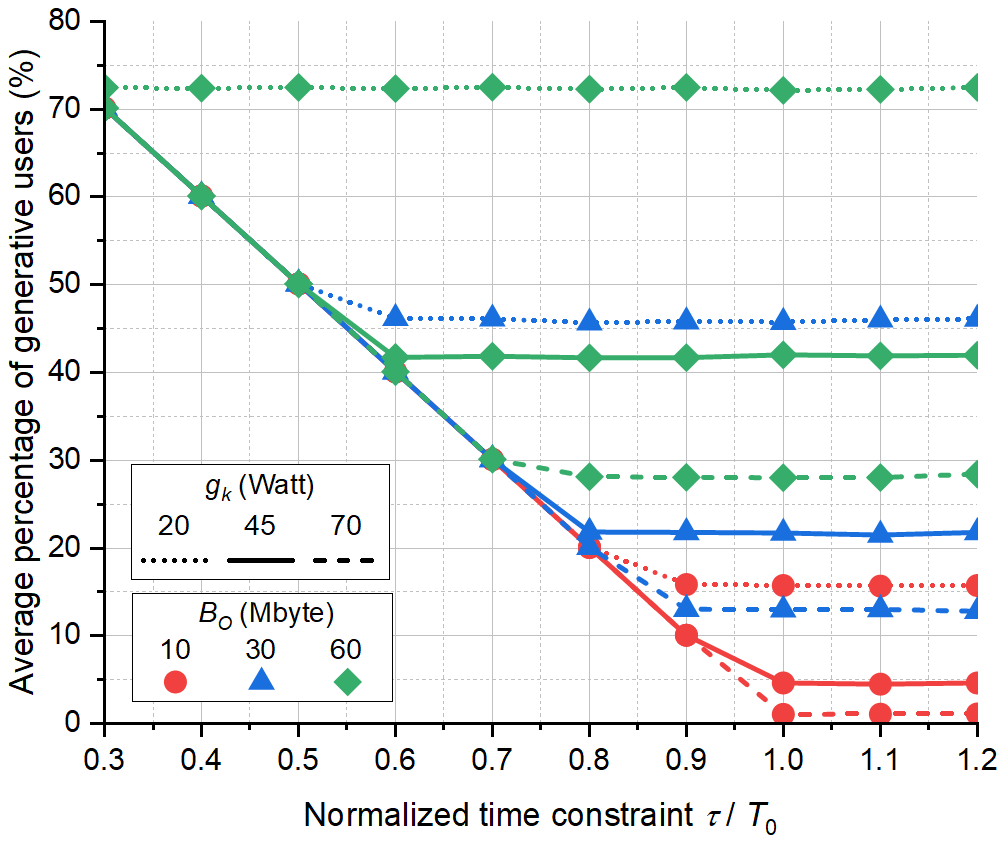}
         \caption{Average ratio of generative users with respect to different time thresholds ($\tau$) for different computation power consumptions ($g_k$) with $B_P=0.1$ Mbyte, where $T_0$ denotes the total transmission time with all non-generative users.}
         \label{Fig:Percentage-of-users_2}
\end{figure}

Fig.~\ref{Fig:Average_beta} shows the expected value of channel power gain $\beta_k$, E[$\beta_k$] in order to investigate an impact of channel power gain $\beta_k$ on generative user selection when $B_O=10, 30, 60$ Mbyte, $g_k=45$ W, and $\tau=T_0$.
Since a channel inversion power control scheme is assumed, users with small $\beta_k$ values will experience severe fading and thus will consume high transmission energy.
Therefore, 
users with favorable $\beta_k$ (i.e., large $\beta_k$) are selected as non-generative users, 
while users with unfavorable $\beta_k$ (i.e., small $\beta_k$) are assigned as generative users in order to reduce the total energy.
The results in Fig.~\ref{Fig:Average_beta} illustrates this, as E[$\beta_k$] of generative users is much smaller than that of non-generative users for all investigated ranges of $B_P$.  
As $B_P$ is increased, portion of generative users will be decreased  
as shown in Fig.~\ref{Fig:Percentage-of-users_1}, and this leads to reduction of E[$\beta_k$]s for not only generative but also non-generative users. 
If the portion of generative users is small, the system will select generative users in much worse channel conditions,
and thus, E[$\beta_k|\text{generative users}$] will be decreased.
In this case, the users who were selected as generative users because of their intermediate channel conditions, will now be moved to non-generative users, 
and E[$\beta_k|\text{non-generative users}$] will also be decreased.
Eventually, if there is no generative users because of too large size of $B_P$, then E[$\beta_k|\text{generative users}$] = 0 because of no users
and E[$\beta_k|\text{non-generative users}$] = 1 because of all users in the system.



Fig.~\ref{Fig:Energy_2} compares the system availability and energy consumption for different time thresholds $\tau$ when $g_k=45$ W and $B_P=0.1$ Mbyte.
As the time constraint is relaxed, the consumed energy also decreases due to the reduction of the interference caused by uplink transmission of generative users. 
In the conventional downloading system (i.e., $n=0$), $\tau$ should be at least $T_0$,
while the proposed method requires much shorter transmission time 
by adjusting the percentage of generative users. 
For example, the proposed scheme shows 59\% of reduction in transmission time to transmit an image of $B_O=60$ Mbyte,
requiring transmission time of $0.4T_0$.

Fig.~\ref{Fig:Percentage-of-users_2} shows the average percentage of generative users
for different time thresholds $\tau$ and consumed computation power $g_k$ when $B_P=0.1$ Mbyte.
As observed in Fig.~\ref{Fig:Percentage-of-users_2}, a higher computation energy $g_k$ results in a lower portion of generative users in order to avoid the burden of computation.
Meanwhile, if the time resource becomes scarce, the percentage of generative users
is increased to meet the time constraint rather than to minimize energy consumption regardless of $B_O$. 
For example, when $\tau=0.3T_0$, about 72\% of users participate image generation task in all the investigated cases.
On the other hand, the percentage of generative users decreases and saturates as $\tau$ becomes sufficient. Especially for large $B_O$, the portion of generative users can easily remain constant 
even if the time resource is enough to transmit images without generative user. 
We would expect almost the same results as in Fig.~\ref{Fig:Percentage-of-users_2} even if we change $B_P$.
This is because changing $B_P$ hardly affect the percentage of generative users as also observed in Figs.~\ref{Fig:Engergy_1} and \ref{Fig:Percentage-of-users_1} when $g_k=45$ W.


\section{Conclusion}
In this paper, we studied a downlink SGC system with users equipped with T2I generators. We formulated a total energy minimization problem, and devised a generative user selection algorithm by converting the original formulation into computationally-efficient multiple binary integer linear programming
problems. Various numerical results demonstrated that the proposed solution 
shows more than 50\% of reduction in total energy consumption, compared to the conventional non-generative downloading scheme. Leveraging its energy efficiency, applying the SGC framework to energy-limited applications could be an interesting topic for future study.

\
\vfill
\pagebreak

\bibliographystyle{ieeetr}
\bibliography{ref}

\vfill
\end{document}